\documentclass[twoside,11pt]{article}

\usepackage[preprint]{jmlr2e}
\usepackage[symbol]{footmisc}
\usepackage{booktabs}
\usepackage{array}
\usepackage{tabularray}
\usepackage{pifont}  
\usepackage{enumitem}
\usepackage{tikz}
\usetikzlibrary{shapes.multipart}
\usetikzlibrary{arrows, positioning, calc}
\usepackage{caption}
\usepackage{subcaption}
\usepackage{listings}
\usepackage{xcolor}
\PassOptionsToPackage{hyphens}{url}\usepackage{hyperref}

\newcommand{\dtaianomaly}{\texttt{dtaianomaly}}

\newcommand{\PreserveBackslash}[1]{\let\temp=\\#1\let\\=\temp}
\newcolumntype{C}[1]{>{\PreserveBackslash\centering}p{#1}}
\newcolumntype{R}[1]{>{\PreserveBackslash\raggedleft}p{#1}}
\newcolumntype{L}[1]{>{\PreserveBackslash\raggedright}p{#1}}
\newcolumntype{P}[1]{>{\centering\arraybackslash}p{#1}}
\newcolumntype{M}[1]{>{\centering\arraybackslash}m{#1}}

\newcommand{\xmark}[1][0.8ex]{\tikz 
    \draw (0,0) circle (#1);
}
\newcommand\pmark[1][0.8ex]{\tikz{
    \filldraw[fill=black] (0,0) -- (0, #1) arc [start angle=90, end angle=270, radius=#1] -- cycle; 
    \draw (0,0) circle (#1);
    }
}
\newcommand\cmark[1][0.8ex]{\tikz 
    \fill (0,0) circle (#1);
}
        
\newcommand\minorsection[1]{\noindent\textbf{#1}\hspace{0.3cm}}

\definecolor{codecomment}{RGB}{61, 122, 122}
\definecolor{codenumbers}{RGB}{128, 128, 128}
\definecolor{codestring}{RGB}{186, 33, 33}
\definecolor{codekeyword}{RGB}{0, 128, 0}
\definecolor{codebackground}{RGB}{242, 242, 242}
\lstdefinestyle{python}{
    backgroundcolor=\color{codebackground},
    commentstyle=\color{codecomment},
    keywordstyle=\color{codekeyword},
    numberstyle=\tiny\color{codenumbers},
    stringstyle=\color{codestring},
    basicstyle=\ttfamily\scriptsize,
    breakatwhitespace=false,
    breaklines=true,
    captionpos=b,
    keepspaces=true,
    numbers=left,
    numbersep=5pt,
    showspaces=false,
    showstringspaces=false,
    showtabs=false,
    tabsize=4
}
\usepackage{lastpage}
\jmlrheading{volume}{2025}{1-\pageref{LastPage}}{data-submitted; Revised date-revised}{date-published}{paper-id}{Louis Carpentier, Wannes Meert, Mathias Verbeke}

\ShortHeadings
    {dtaianomaly: A Python library for time series anomaly detection}
    {Carpentier \textit{et al.}}
\firstpageno{1}

\begin{document}

\title{
    dtaianomaly \\ 
    A Python library for time series anomaly detection
}

\author{%
    \name Louis Carpentier\textsuperscript{1,2,3} \email louis.carpentier@kuleuven.be \\
    \name Nick Seeuws\textsuperscript{2} \email nick.seeuws@kuleuven.be \\
    \name Wannes Meert\textsuperscript{1,2} \email wannes.meert@kuleuven.be \\
    \name Mathias Verbeke\textsuperscript{1,2,3} \email mathias.verbeke@kuleuven.be \\
    \addr
    \textsuperscript{1}KU Leuven, Department of Computer Science, Belgium \\
    \textsuperscript{2}Leuven.AI - KU Leuven Institute for AI, Belgium \\
    \textsuperscript{3}Flanders Make@KU Leuven \\
}
\editor{My editor}

\maketitle

\begin{abstract}
    \dtaianomaly~is an open-source Python library for time series anomaly detection, designed to bridge the gap between academic research and real-world applications. Our goal is to (1)~accelerate the development of novel state-of-the-art anomaly detection techniques through simple extensibility; (2)~offer functionality for large-scale experimental validation; and thereby (3)~bring cutting-edge research to business and industry through a standardized API, similar to \texttt{scikit-learn} to lower the entry barrier for both new and experienced users. Besides these key features, \dtaianomaly~offers (1)~a broad range of built-in anomaly detectors, (2)~support for time series preprocessing, (3)~tools for visual analysis, (4)~confidence prediction of anomaly scores, (5)~runtime and memory profiling, (6)~comprehensive documentation, and (7)~cross-platform unit testing. 
    
    The source code of \dtaianomaly, documentation, code examples and installation guides are publicly available at \url{https://github.com/ML-KULeuven/dtaianomaly}. 
\end{abstract}


\section{Introduction}

Time series anomaly detection (TSAD) is the task of identifying anomalous or strange observations in time series data. Accurate detection of anomalies in time series data is crucial for numerous real-world applications~\citep{ukil2016iot}. Due to its significance, TSAD is an active research area, with dozens of new algorithms proposed each year~\citep{liu2024elephant, schmidl2022anomaly}. However, many of these methods are not readily applicable to industrial use cases out of the box. 

\dtaianomaly~is a Python library for time series anomaly detection, which is designed to bridge the gap between cutting-edge research and real-world applications in business and industry. Existing tools for TSAD often have several shortcomings. TimeEval~\citep{WenigEtAl2022TimeEval} is an extensive benchmarking tool, but requires additional setup beyond simple Python scripts to apply the anomaly detectors because they are provided via Docker containers. Aeon~\citep{aeon24jmlr} is a Python library for time series machine learning in general, but only offers experimental support for anomaly detection. TSB-AD~\citep{liu2024elephant} provides standalone implementations for many state-of-the-art anomaly detection algorithms but lacks a cohesive framework. Furthermore, the absence of clear, user-friendly documentation makes it more difficult for new users to understand and adopt the tool.

To effectively bring cutting-edge anomaly detection research to real-world applications, \dtaianomaly~targets both researchers developing novel anomaly detection techniques and industry practitioners leveraging these methods for operational decision-making. This is achieved by designing \dtaianomaly~with the following key principles in mind (click the URLs to find more information and code examples in the documentation): 
\begin{enumerate}[noitemsep,leftmargin=0.5cm]
    \item \textbf{Standardized API.} All models in \dtaianomaly~adhere to a consistent interface, inspired by \texttt{scikit-learn}~\citep{pedregosa2011scikit} and \texttt{Pyod}~\citep{zhao2019pyod}. This ensures a minimal learning curve for new users while providing a familiar framework for experienced developers. \\
    {\scriptsize \url{https://dtaianomaly.readthedocs.io/en/stable/getting\_started/examples/anomaly_detection.html}}

    \item \textbf{Extensibility.} The base objects in \dtaianomaly~are designed for easy extension. Users only need to implement up to two core methods, such as \texttt{.fit()} and \texttt{.predict()}, enabling rapid development of new state-of-the-art anomaly detection models and seamless integration in real-world use cases. \\
    {\scriptsize \url{https://dtaianomaly.readthedocs.io/en/stable/getting_started/examples/custom_models.html}}

    \item \textbf{Comprehensive experimental validation.} \dtaianomaly~simplifies quantitative evaluation by allowing users to benchmark multiple anomaly detectors, including custom models, with just a few lines of code. This approach streamlines experimental setups and enhances reproducibility, but also offers a robust manner of identifying the best model for the given use case. Additionally, \dtaianomaly~provides visualization tools for qualitative assessments of model performance. \\
    {\scriptsize \url{https://dtaianomaly.readthedocs.io/en/stable/getting_started/examples/quantitative_evaluation.html}}
\end{enumerate}

\begin{table}
    \centering
    \newlength{\widthpackagecolumn}
    \setlength{\widthpackagecolumn}{2.25cm}
    \caption{Comparison of \dtaianomaly~with existing time series anomaly detection tools.}
    \begin{tabular}{L{3.8cm} | *{3}{C{\widthpackagecolumn}} | C{\widthpackagecolumn}}
        \toprule
        ~ & TimeEval & aeon & TSB-AD &  \dtaianomaly \\            
        \midrule
        Statistical models     & \cmark & \cmark & \cmark & \cmark \\
        Neural models          & \cmark & \xmark & \cmark & \pmark \\
        Foundation models      & \xmark & \xmark & \cmark & \pmark \\
        Preprocessing          & \pmark & \cmark & \pmark & \cmark \\
        Visualization          & \xmark & \cmark & \pmark & \cmark \\
        Confidence prediction  & \xmark & \xmark & \xmark & \cmark \\
        Memory profiling       & \pmark & \xmark & \xmark & \cmark \\
        Documentation          & \cmark & \cmark & \pmark & \cmark \\
        Unit testing           & \cmark & \cmark & \xmark & \cmark \\
        \bottomrule
        \multicolumn{5}{l}{
            \footnotesize
            \hspace{1cm} 
            \cmark[0.6ex]exists in tool; 
            \pmark[0.6ex]partially exists in tool; 
            \xmark[0.6ex]does not exist in tool.
        }
    \end{tabular}
    \label{tab:comparison-packages}
\end{table}

In addition to these three key principles, \dtaianomaly~offers several features, as listed in Table~\ref{tab:comparison-packages}. (1)~\dtaianomaly~supports a broad range of anomaly detection algorithms.\footnote{Implementations of neural and foundation models are available upon request, but have not yet been integrated into the main code base as they are still experimental.} (2)~\dtaianomaly~provides a wide range of time series preprocessing methods, such as scaling, resampling, and moving averages. (3)~\dtaianomaly~includes multiple visualization techniques to facilitate qualitative analysis and offer a more intuitive understanding of the model's behavior. (4) Beyond anomaly score prediction, \dtaianomaly~is able to quantify the confidence of individual predictions~\citep{perini2021quantifying}. (5)~\dtaianomaly~supports profiling both runtime and memory consumption of anomaly detectors, making it easier to assess their feasibility for deployment on edge devices~\citep{murshed2022machine}. (6)~\dtaianomaly~offers comprehensive documentation of its standardized API, along with multiple code examples to lower the entry barrier for new users. (7)~\dtaianomaly~includes rigorous unit testing and cross-platform continuous integration to ensure the correctness and robustness of its implementations.

\section{Library Design and Implementation}

\begin{figure}[bh]
    \centering
    \definecolor{ku-leuven-blue}{RGB}{220, 231,240}
    \definecolor{ku-leuven-dark-blue}{RGB}{47, 77, 93}
    \definecolor{ku-leuven-medium-blue}{RGB}{29, 141, 176}
    \definecolor{data}{RGB}{241, 208, 135}
    \definecolor{preprocessor}{RGB}{231, 176, 55}
    \definecolor{detector}{RGB}{192, 139, 23}
    \definecolor{thresholding}{RGB}{229, 232, 142}
    \definecolor{metric}{RGB}{172, 175, 36}
    \definecolor{visualization}{RGB}{114, 117, 24}
    \begin{subfigure}[b]{0.47\textwidth}
         \centering
        \begin{tikzpicture}[
            node distance=2cm,
            every node/.style={draw, rectangle, minimum height=0.8cm, minimum width=3cm, rounded corners, align=center, thick}
        ]
            \node[fill=data] (data) at (0.0, 0.0) {\texttt{LazyDataLoader}};
            \node[fill=ku-leuven-medium-blue, below=0.4cm of data] (pipeline) {
            \textcolor{white}{\texttt{Pipeline}} \\
                \begin{tikzpicture}
                    \node[fill=preprocessor] (preprocessor) at (-1.8, 0) {\texttt{Preprocessor}};
                    \node[fill=detector] (detector) at (1.8, 0) {\texttt{BaseDetector}};
                    \draw[-{latex}, thick] (preprocessor) -- (detector);
                \end{tikzpicture}
            };
            \node[circle, fill, draw, inner sep=0.7, minimum width=0cm, minimum height=0cm] (dot) at (0, -2.5) {};

            \node[fill=metric] (probametric) at (1.8, -3.2) {\texttt{ProbaMetric}};
            \node[fill=thresholding] (thresholding) at (-1.8, -3.2) {\texttt{Thresholding}};
            \node[fill=metric] (binarymetric) at (-1.8, -4.4){\texttt{BinaryMetric}};
            \node[fill=visualization] at (1.8, -4.4){\texttt{Visualization}};

            \draw[-{latex}, thick] (data) -- (pipeline);
            \draw[thick] (pipeline) -- (dot);
            \draw[-{latex}, thick] (dot) -- (probametric.north west);
            \draw[-{latex}, thick] (dot) -- (thresholding.north east);
            \draw[-{latex}, thick] (thresholding) -- (binarymetric);
        \end{tikzpicture}
        \caption{The modules and their interaction.}
        \label{fig:modules}
     \end{subfigure}
     \hfill
     \begin{subfigure}[b]{0.47\textwidth}
        \centering
        \begin{tikzpicture}[
            node distance=1.5cm,
            every node/.style={draw, rectangle, minimum height=1cm, minimum width=2.6cm, rounded corners, align=center, thick}
        ]
            \node[fill=ku-leuven-dark-blue, text=white] (data) at (0.0, 0.0) {Datasets};
            \node[below of=data, fill=ku-leuven-dark-blue, text=white] (preprocessors) {Preprocessors};
            \node[below of=preprocessors, fill=ku-leuven-dark-blue, text=white] (detectors) {Detectors};
            \node[below of=detectors, fill=ku-leuven-dark-blue, text=white] (metrics) {Metrics};

            \draw[-{latex}, thick] (data.east) -- ++(0.45, 0); 
            \draw[-{latex}, thick] (preprocessors.east) -- ++(0.45, 0);
            \draw[-{latex}, thick] (detectors.east) -- ++(0.45, 0);
            \draw[-{latex}, thick] (metrics.east) -- ++(0.45, 0);
            
            \node[minimum height=5.5cm, minimum width=4cm, fill=ku-leuven-blue] (workflow) at (3.75,-2.25) {
            \texttt{Workflow} \\
                \begin{tikzpicture}[node distance=3.2cm]
                    \tikzstyle{mycircle} = [circle, minimum width=0.9cm, minimum height=0.9cm]
                    \tikzstyle{myblock} = [minimum height=0.6cm, minimum width=3.3cm]
                    
                    \node[draw=none] at (0, 1.4) {}; 

                    \node[mycircle, fill=data] (datacircle) at (-1.2, 1.2) {};
                    \node[mycircle, fill=preprocessor] (preprocessorcircle) at (0, 1.2) {};
                    \node[mycircle, fill=detector] (detectorcircle) at (1.2, 1.2) {};

                    \node[myblock, fill=data] at (0, 0) {Time series};
                    \node[myblock, fill=preprocessor] at (0, -0.6) {Preprocessor};
                    \node[myblock, fill=detector] at (0, -1.2) {Anomaly detector};

                    \node[mycircle, fill=metric] (metriccircle) at (0, -2.4) {};
                    
                    \draw[-{latex}, thick] (datacircle.south) -- ++(0, -0.45); 
                    \draw[-{latex}, thick] (preprocessorcircle.south) -- ++(0, -0.45); 
                    \draw[-{latex}, thick] (detectorcircle.south) -- ++(0, -0.45); 
                    \draw[{latex}-, thick] (metriccircle.north) -- ++(0, 0.45); 

                    \tikzstyle{iconnode} = [draw=none, minimum height=0cm, minimum width=0cm]
                    \node[iconnode] at (datacircle) {\includegraphics[width=0.6cm]{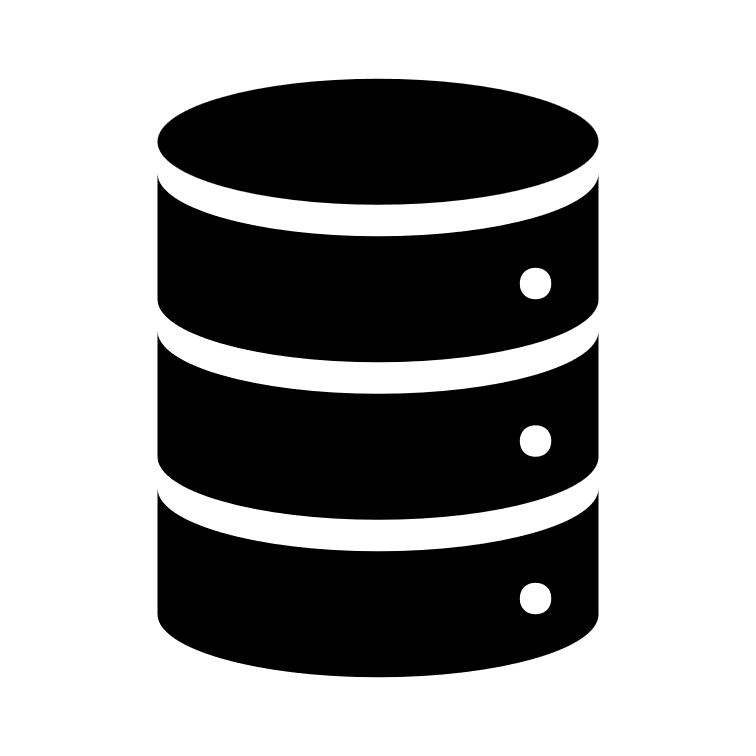}};
                    \node[iconnode] at (preprocessorcircle) {\includegraphics[width=0.6cm]{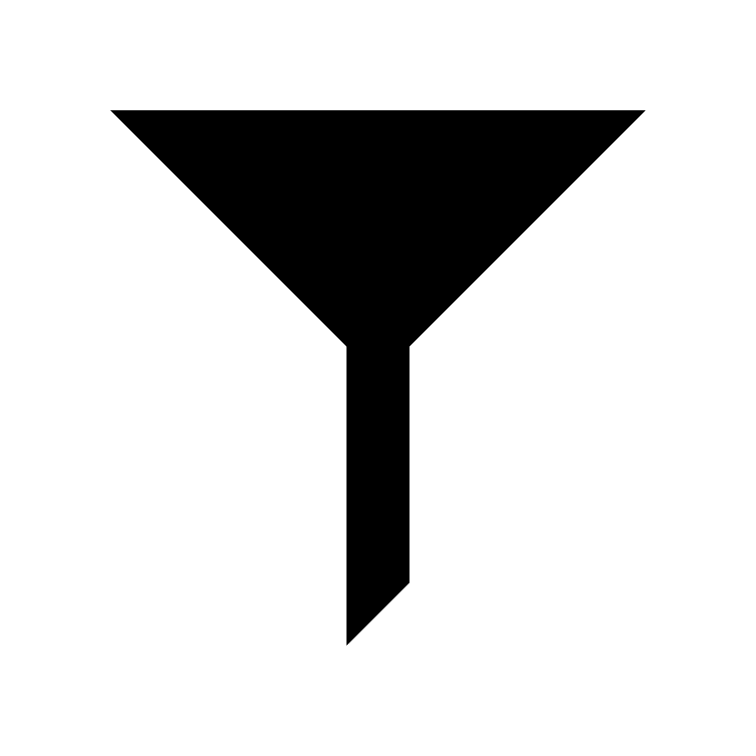}};
                    \node[iconnode] at (detectorcircle) {\includegraphics[width=0.6cm]{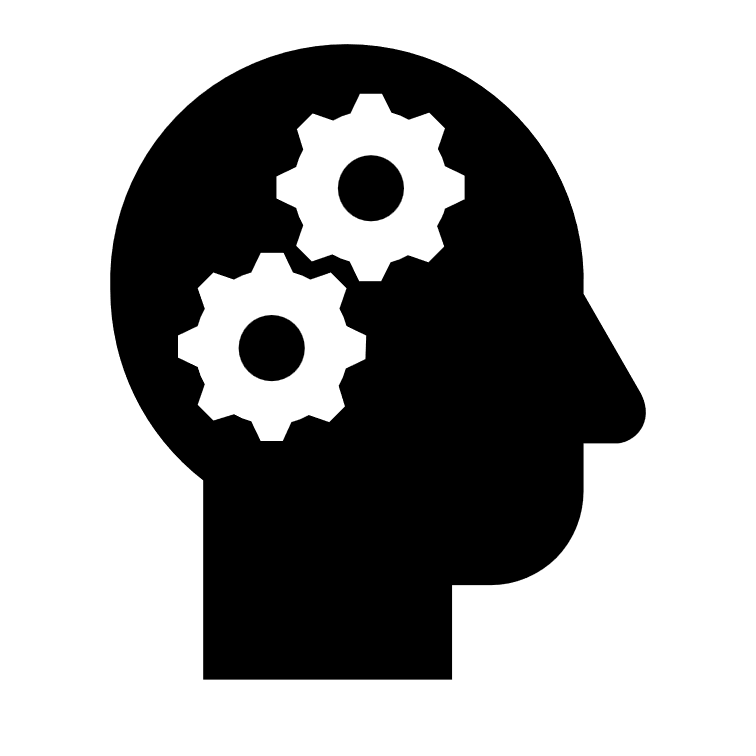}};
                    \node[iconnode] at (metriccircle) {\includegraphics[width=0.6cm]{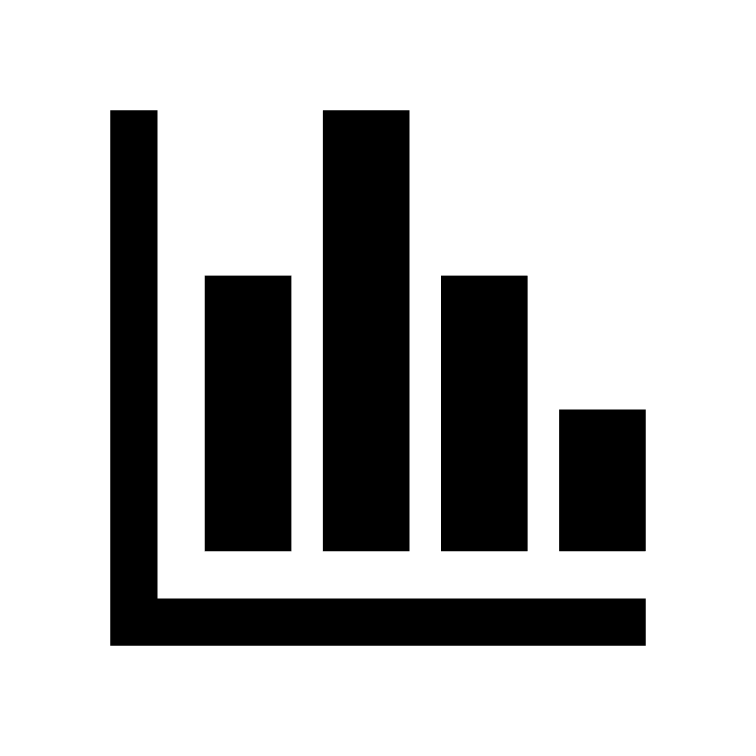}};
                \end{tikzpicture}
            };
        \end{tikzpicture}
        \caption{The \texttt{Workflow}.}
        \label{fig:workflow}
    \end{subfigure}
    \caption{The base objects in \dtaianomaly~and their interactions.}
    \label{fig:dtaianomaly-design}     
\end{figure}
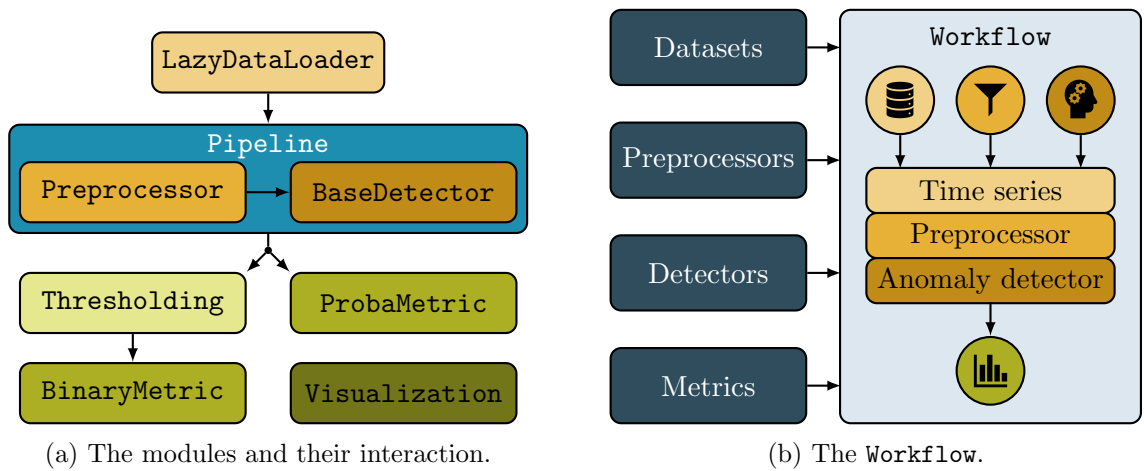

\dtaianomaly~adopts an object-oriented design similar to \texttt{scikit-learn}. The only exception is the \texttt{visualization} module, which follows a functional design. The library organizes its components into distinct modules based on their functionality, ensuring a consistent inheritance structure where a base class defines the interface for each module. Figure~\ref{fig:dtaianomaly-design} visualizes the core components in \dtaianomaly~and their interactions. 

The main component is the \texttt{BaseDetector}, a standardized interface implemented by all anomaly detectors in \dtaianomaly. The models are initialized and trained on training data using the \texttt{.fit(X, y)} method.\footnote{The \texttt{y}-parameter in \texttt{.fit(X, y)} is optional, as anomaly detection is generally unsupervised; it is included for consistency.} Once the model is fitted, there are two methods to detect anomalies in a new, unseen time series: (1)~\texttt{.decision\_function(X)} computes the raw anomaly scores which depend on the model (e.g., distances for Matrix Profile), and (2)~\texttt{.predict\_proba(X)} generates the probability of an observation being anomalous.

Figure~\ref{fig:modules} illustrates the typical TSAD loop using \dtaianomaly~components. The process begins with loading the time series via a \texttt{LazyDataLoader}, which employs just-in-time loading to prevent unnecessary storage of large datasets. However, \dtaianomaly~also supports caching to avoid redundant loading of the same data. Before anomaly detection, time series data is typically preprocessed using one or more \texttt{Preprocessor}s. A \texttt{BaseDetector} can then be used to detect anomalies in the processed time series. \dtaianomaly~also allows users to chain multiple preprocessing steps with an anomaly detection model into a single \texttt{Pipeline} object, enabling seamless execution of both steps through a single function call. 

To quantitatively evaluate anomaly detection performance, \dtaianomaly~provides multiple metrics. Users can apply \texttt{ProbaMetric}s, such as Area Under the Curve (AUC) metrics, which take continuous anomaly scores as input. Alternatively, \texttt{BinaryMetric}s, such as precision and recall, require \texttt{Thresholding} to convert continuous decision scores into binary anomaly labels. Additionally, the \texttt{visualization} module offers various methods for qualitatively assessing model performance. A comprehensive and up-to-date overview of implemented visualization methods is available at \url{https://dtaianomaly.readthedocs.io/en/stable/api/visualization.html}. A practical example of anomaly detection using \dtaianomaly~is given in Section~\ref{sec:illustrative-example}.

Beyond single time series anomaly detection, \dtaianomaly~facilitates large-scale experimental validation through the \texttt{Workflow} module, illustrated in Figure~\ref{fig:workflow}. Users can define lists of \texttt{LazyDataLoader}s, \texttt{Preprocessor}s, \texttt{BaseDetector}s, and \texttt{Metric}s (optionally including \texttt{Thresholding}s). By invoking the \texttt{.run()} method, the \texttt{Workflow} systematically evaluates all model-dataset combinations in a grid-like manner, computing evaluation metrics along with runtime and memory usage. Results are returned as a \texttt{pandas.DataFrame} for easy analysis. By automating experiment execution, performance evaluation, and error logging, the \texttt{Workflow} enables large-scale validation of anomaly detection models without user intervention.

\section{Illustrative example of \dtaianomaly} \label{sec:illustrative-example}

The code example below demonstrates how to detect anomalies using \dtaianomaly~with just three lines of code. First, the time series is loaded. Next, the anomaly detector is trained, during which the window size is automatically determined based on the dominant Fourier frequency~\citep{ermshaus2023window}. Finally, the anomaly scores are predicted.\footnote{Notice that \dtaianomaly~utilizes the \texttt{.decision\_function()} method to detect anomalies instead of the more common \texttt{.predict()} method. This choice is made because anomaly scores are not binary predictions, as expected for \texttt{.predict()}, but rather continuous values. For example, they represent z-normalized Euclidean distances in the case of Matrix Profile.}

The time series consists of normal heartbeats, PVC beats, and wandering baselines, with one beat replaced by a constant section of very low amplitude noise~\citep{wu2023current}. Detecting anomalies in this time series, which consists of 45,000 observations, takes less than two seconds. For comparison, it takes only 8 seconds to detect anomalies in \texttt{200\_UCR\_Anomaly\_tiltAPB4\_20000\_67995\_67996}, which has over 105,000 observations.

\begin{lstlisting}[language=Python]
from dtaianomaly.data import UCRLoader
from dtaianomaly.anomaly_detection import MatrixProfileDetector

# Load the data, fit on the train set, and predict on the test set
data = UCRLoader('002_UCR_Anomaly_DISTORTED2sddb40_35000_56600_56900.txt').load()
detector = MatrixProfileDetector(window_size='fft').fit(data.X_train)
y_pred = detector.decision_function(data.X_test)
\end{lstlisting}

\begin{figure}[h]
    \centering
    \includegraphics[width=\linewidth]{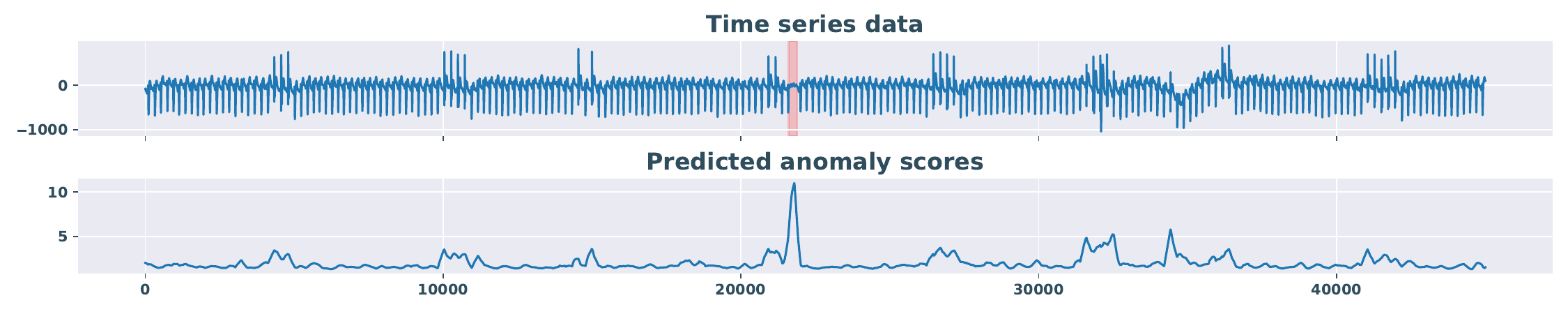}
    \caption{The predicted anomaly scores, visualized using \href{https://dtaianomaly.readthedocs.io/en/stable/api/visualization.html\#dtaianomaly.visualization.plot\_anomaly\_scores}{\texttt{plot\_anomaly\_scores}}.}
    \label{fig:example}    
\end{figure}

\section{Conclusions and future plans}

In this work, we introduced \dtaianomaly, a publicly available Python library to bring state-of-the-art time series anomaly detection to real-world use cases in business and industry. Its key features include (1) a standardized API, (2) high extensibility, and (3) the ability to conduct comprehensive experimental evaluations. Moving forward, we plan to expand support for more state-of-the-art anomaly detection methods. Additionally, we will focus on developing generic wrapper approaches for existing detectors, leveraging the standardized API to enhance the capabilities of time series anomaly detection. Our efforts will concentrate on the following key areas:

\minorsection{Context-aware anomaly detection}
In real-world scenarios, a wealth of contextual information is available (e.g., time of day, season, temperature). However, most anomaly detectors disregard this context and analyze only the raw data, despite the fact that \emph{``normality"} heavily relies on context~\citep{thill2017time}. Our goal is to develop techniques that enhance existing anomaly detectors by incorporating contextual information, enabling them to become context-aware.

\minorsection{Human-in-the-loop}
While anomaly detection is traditionally treated as an unsupervised problem, many practical applications benefit from expert feedback. We plan to develop strategies to incorporate feedback from domain experts into the anomaly detection process, improving both model performance and adaptability through interactive feedback loops.

\minorsection{Automatic anomaly detector selection}
Given the vast number of anomaly detection methods available, selecting the most suitable one can be challenging. Moreover, many detectors are highly sensitive to their hyperparameters. To address these challenges, we will develop techniques to automatically select the optimal anomaly detectors and corresponding hyperparameters for a given time series, thereby greatly enhancing the usability of \dtaianomaly.

\acks{%
    This research is supported by Flanders Innovation \& Entrepreneurship (VLAIO) through the AI-ICON project CONSCIOUS (HBC.2020.2795), the Flemish government under the Flanders AI Research Program, and by Internal Funds of KU Leuven (STG/21/057).
}

\vskip 0.2in
\bibliography{bibliography}

\end{document}